\documentclass[letterpaper, 10 pt, conference]{ieeeconf}  

\IEEEoverridecommandlockouts                              

\usepackage{import}
\usepackage[pdftex]{graphicx}
\usepackage{sidecap}
\usepackage{cite}
\usepackage{verbatim}		
\usepackage{amsmath}
\usepackage{amssymb}

\usepackage{amsthm}
\usepackage{mathtools}	
\usepackage{grffile}	
\usepackage[tight,footnotesize]{subfigure}
\usepackage{microtype} 
\usepackage{color}
\usepackage{url}
\usepackage[ruled,vlined,linesnumbered]{algorithm2e}
\usepackage{bm} 
\usepackage{comment}
\usepackage{arydshln}
\let\labelindent\relax
\usepackage{enumitem}

\usepackage{adjustbox}
\usepackage{tikz}
\usetikzlibrary{shadings, patterns, angles, quotes, arrows.meta, shapes, decorations.pathmorphing, decorations.shapes, decorations.text, automata, positioning}
\usetikzlibrary{calc,intersections,arrows.meta}
\usepackage{pgfplots}

	\tikzset{
	pil/.style={
		->,
		thick,
		shorten <=2pt,
		shorten >=2pt,}
}

\theoremstyle{remark}
\newtheorem{remarkx}{Remark}
\newenvironment{remark}
{\pushQED{\qed}\remarkx}
{\popQED\endremarkx}
\newtheorem{examplex}{Example}
\newenvironment{example}
{\pushQED{\qed}\examplex}
{\popQED\endexamplex}

\theoremstyle{definition}
\newtheorem{defn}{Definition}
\newtheorem{assump}{Assumption}
\newtheorem{problem}{Problem}
\newtheorem*{problem*}{Problem}

\theoremstyle{plain}
\newtheorem{theorem}{Theorem}
\newtheorem{lemma}{Lemma}

\newtheorem{prop}{Proposition}

\newcommand{\cmmnt}[1]{}

\usepackage[hidelinks,linktoc=all]{hyperref}

\title{\LARGE \bf
Inverse Kinematics on Guiding Vector Fields for Robot Path Following
}

\author{Yu Zhou*, Jesús Bautista*, Weijia Yao, Héctor García de Marina 
	\thanks{Yu Zhou is with INRIA at Lille, France. J. Bautista and H.G. de Marina are with the Department of Computer Engineering, Automation, and Robotics, University of Granada (UGR-CITIC), Spain. Weijia Yao is with Hunan University, China. This work is supported by the ERC Starting Grant \emph{iSwarm} 101076091 and the RYC2020-030090-I grant from the Spanish Ministry of Science. *Yu and Jesús contributed equally to this work.}
}

\begin{document}

\maketitle
\thispagestyle{empty}
\pagestyle{empty}

\begin{abstract}
Inverse kinematics is a fundamental technique for motion and positioning control in robotics, typically applied to end-effectors. In this paper, we extend the concept of inverse kinematics to guiding vector fields for path following in autonomous mobile robots. The desired path is defined by its implicit equation, i.e., by a collection of points belonging to one or more zero-level sets. These level sets serve as a reference to construct an error signal that drives the guiding vector field toward the desired path, enabling the robot to converge and travel along the path by following such a vector field. We start with the formal exposition on how inverse kinematics can be applied to guiding vector fields for single-integrator robots in an $m$-dimensional Euclidean space. Then, we leverage inverse kinematics to ensure that the level-set error signal behaves as a linear system, facilitating control over the robot's transient motion toward the desired path and allowing for the injection of feed-forward signals to induce precise motion behavior along the path. We then propose solutions to the theoretical and practical challenges of applying this technique to unicycles with constant speeds to follow 2D paths with precise transient control. We finish by validating the predicted theoretical results through real flights with fixed-wing drones.
\end{abstract}

\section{Introduction}
In robots with joint actuators and end-effectors, inverse kinematics utilizes the kinematic equations that relate them to determine the joint parameters required to achieve a desired configuration, such as the position and orientation of each end-effector \cite{paul1981robot,choset2005principles}. The sensitivity of the end-effector configuration concerning small variations of the joints is determine by a Jacobian matrix, and the inverse kinematics typically consists in taking the Jacobian's pseudo-inverse to calculate the actuation to induce the desired motion on the end-effectors. On the other hand, recent advances in \emph{Guiding Vector Fields} (GVFs) allow autonomous mobile robots with non-holonomic constraints to follow paths without the risk of encountering \emph{singularities}, where the robot can get stuck. This risk can be eliminated globally with a parametric description of the path \cite{yao2021:Tro}, or locally if the Jacobian associated with the implicit equation of the path does not vanish along the desired path \cite{kapitanyuk2017:TCST}, e.g., for open or non-self-intersecting paths. One consequence of describing paths with a parametric or implicit equation is the impact on the dimension of the error signal. For example, for a 2D path, the parametric description requires a three-dimensional error signal (two Cartesian plus one virtual) as shown in \cite{yao2021:Tro}, while the implicit representation of the path requires only a scalar error signal \cite{kapitanyuk2017:TCST}. To be concise, we will focus on the latter in this paper.

In scenarios where the desired path is defined by implicit functions, it is natural to consider a vector field with a \emph{convergence term} that aligns with the gradients of these functions. This ensures that the system is quickly attracted to the zero-level sets \cite{lawrence2007:GNCC, goncalves2010:Tro, kapitanyuk2017:TCST,yao2021:Tro}. One of the key advantages of the vector field approach is its flexibility, as the vector field does not necessarily need to follow the gradient direction but just \emph{ascending}. Exploiting this fact, this paper presents the application of inverse kinematics to guiding vector fields, enabling the error signal, based on level set values, to behave as a designed system such as a linear one. This technique allows for precise control during the transitory motion of the robot toward the desired path and can even enable feed-forward behavior around such a path. This is an aspect that has not been explored in previous works \cite{lawrence2007:GNCC, goncalves2010:Tro, yao2021:Tro, kapitanyuk2017:TCST, de2017guidance}. We will demonstrate the effectiveness of this technique on single integrator robots in $m$-dimensions and on 2D unicycles with constant speeds.

One of the motivations for this work arises from the need to improve telemetry reception from small fixed-wing drones. Typical onboard dipole antennas have a toroidal radiation pattern. When the drone is far from the base station, the dipole might radiate primarily toward the immediate ground due to the bank angle required to follow a curved path, leading to a loss of communication with the base station during certain, possibly long, segments of the flight. While directional antennas with active tracking at the base station can mitigate this issue, for smaller drones where every gram counts, an additional solution is to induce an oscillatory motion around the path. Consequently, oscillations in the bank angle can improve telemetry reception when using dipole antennas onboard. Of course, while directly injecting an arbitrary oscillatory signal into the roll angle might seem like a straightforward solution, it risks causing divergence from the desired path, as it would not be in a closed-loop system with any path-error signal. This is why we exploit the inverse kinematics to inject such oscillation but into the level-set error signal. This approach not only ensures that the oscillation has a clear interpretation in length units but, as anticipated, it also allows for precise control of the transitory motion towards and around the desired path.

Compared to other techniques in the literature, this problem could be approached using \textit{Model Predictive Control} (MPC), \textit{Control Barrier Functions} (CBFs), or other optimization-based methods that may incorporate telemetry reception constraints. However, these techniques require a numerical solver in general while our proposed solution is purely reactive with an analytical expression.

The article is organized as follows. Section \ref{sec: vf} introduces notation, defines guiding vector fields, discusses necessary and sufficient conditions, and formally states the problem. In Section \ref{sec: ik}, inverse kinematics is applied to the error signal, ensuring a singularity-free vector field. Section \ref{sec: uni} adapts the vector field for 2D unicycles with constant speeds, introducing a tangential component that preserves velocity without affecting the error signal. Fixed-wing drone experiments are presented in Section \ref{sec: sim}, demonstrating effectiveness, followed by conclusions in Section \ref{sec: con}.

\section{Guiding vector field for path following and problem statement}
\label{sec: vf}
Consider $p \in \mathbb{R}^m$ as the position of an autonomous mobile robot whose dynamics are the single integrator\footnote{Throughout the paper, we choose the level of verbosity on writing variable dependencies such as from $u_p(\phi(p(t)), t)$ to $u_p$ depending on the context.} $\dot p = u_p, u_p\in\mathbb{R}^m$. We describe the desired path in the $m$-dimensional Euclidean space as the intersection of $m-1$ hypersurfaces
\[
	\mathcal{P}:=\left\{p\in \mathbb{R}^m: \phi_i(p)=0, \, i\in\{1,\dots, m-1\}\right\}.
\]
where $\phi_i: \mathbb{R}^m \to \mathbb{R}$ is at least $m$ times differentiable, i.e., $\phi_i \in C^m$. For $m = 2$, $\phi(p)$ is just the implicit equation describing the path, e.g., $\phi(p) = p_x^2 + p_y^2 - r^2$ for the circumference with radius $r$. We just stack all $\phi_i(p)$ to construct our error, i.e., $\phi(p) = \begin{bmatrix} \phi_1(p) & \phi_2(p) & \dots & \phi_{m-1}(p) \end{bmatrix}^\top \in \mathbb{R}^{m-1}$ for the design of the guiding vector fields. Indeed, when $\phi(p)= 0$, the robot is on the desired path; then we can define the distance of the position $p$ to the desired path $\mathcal{P}$ by $d_{\mathcal{P}}(p) := \|\phi(p)\|$. In this paper, we are going to consider only \emph{well-defined paths} by considering two assumptions:
\begin{assump}\label{asp:2}
    $\mathcal{P}$ is nonempty and connected.
\end{assump}
\begin{assump}\label{asp:1}
The gradients $\nabla \phi_i(p)\in\mathbb{R}^m$ for all $i \in\{1,\dots,m-1\}$ are linearly independent at $p \in \Omega_b$, where $\Omega_b := \{p\in\mathbb{R}^m : \|\phi(p)\| \leq b, \, b \in \mathbb{R}^+ \}$ is a neighborhood of the path $\mathcal{P}$.
\end{assump}

Assumption \ref{asp:1} helps fulfill a necessary condition for a guiding vector field by ensuring the transversality of the intersection of all hypersurfaces $\phi_i(p)$, which, by the transversality theorem \cite{golubitsky2012:book}, implies that $\mathcal{P}$ is a smooth 1-dimensional manifold in $\mathbb{R}^m$. Moreover, since the gradients are linearly independent for $p \in \Omega_b$, the Jacobian $J_\phi = \left[\begin{smallmatrix}\nabla \phi_1(p) & \cdots & \nabla \phi_{m-1}(p)\end{smallmatrix}\right]^\top \in \mathbb{R}^{(m-1)\times m}$ is full rank for $p \in \Omega_b$.


The kinematic model for the error signal is straightforward to derive by the chain rule, i.e., 
\begin{equation}\label{eq: sys_phi}
	\dot{\phi}(p(t)) = J_\phi\left(p(t)\right) \, \dot p(t) = J_\phi\left(p(t)\right) \,u_p.
\end{equation}
In the kinematic model (\ref{eq: sys_phi}), the image of $J_\phi(p)$ is known as the \emph{input space}, i.e., the set of all achievable rates of change in $\phi(p)$ through variations in $p$. 
Consider the vector field $f:\mathbb{R}^m\to \mathbb{R}^m$ for the single integrator,
\begin{equation}\label{eq: field_xi}
	\dot p(t) = f\left(\phi(p), t\right),
\end{equation}
and we call it \emph{guiding vector field} if it fits into the following definition:
\begin{defn} \label{def: gvf}The field (\ref{eq: field_xi}) is a \emph{Guiding Vector Field} if:
\begin{enumerate}
	\item The kinematic system described by \eqref{eq: sys_phi} and \eqref{eq: field_xi} is practically (asymptotically) stable with respect to the path $\mathcal{P}$, meaning that for some constant $a > 0, (a = 0)$
$d_{\mathcal{P}}(p(t)) \le \gamma\left(d_{\mathcal{P}}\left(p\left(0\right)\right), t\right) + a, \, \gamma \in \mathcal{KL}, \, p(0) \in \Omega_b$, where $\mathcal{KL}$ is a class of function that describes how the distance decreases over time \cite{khalil}.
\item Consider the manifold $\mathcal{M}{\phi_z}$ defined by the level sets $\phi(p) = z$ for a fixed vector $z \in \mathbb{R}^{m-1}$, and the tangent space $T_p(\mathcal{M}\phi_z)$ for $p\in\mathcal{M}{\phi_z}$. For all the positions $p\in\Omega_b$ we have that $f\left(\phi(p), t\right) \cap T_p(\mathcal{M}\phi_z) \setminus\{0\}\neq \emptyset$, i.e, the vector field ensures \emph{progressing} along the path. 
\end{enumerate}
\end{defn}

The second condition of Definition \ref{def: gvf} can be relaxed by requiring the projection only for $p \in \mathcal{P}$. However, Proposition \ref{prop: jinv} in this section will show that Assumption \ref{asp:1} fits into the second condition of Definition \ref{def: gvf}.

The implicit function theorem \cite{do2016:book, lee2012:book} states that the tangent space $T_p(\mathcal{M}\phi_z)$ at any point $p$ in the level set $\mathcal{M}\phi_z$ lies within the kernel of the Jacobian matrix $J_\phi(p)$, i.e., $T_p(\mathcal{M}\phi_z) \subset \text{Ker}\{J_\phi(p)\}|_{\phi(p)=z}$. Then, we say that a vector $v_T \in \mathbb{R}^m$ belongs to the tangent space $T_p(\mathcal{M}\phi_z)$ at a point $p \in \mathcal{M}\phi_z$ only if it satisfies the condition $J_\phi(p) v_T = 0$. In other words, according to \eqref{eq: sys_phi}, the robot moves with velocity $v_T$ while $\phi(p)$ remains constant and equals to $z$. More formally, since $J_\phi(p)$ is full rank then $T_p(\mathcal{M}{\phi_z}) := \{v_T \in \mathbb{R}^m : J_\phi(p) v_T= 0, \phi(p) = z\}$, whose dimension is $m - (m - 1) = 1$.

\begin{example}
\label{ex: gvf}
We check whether $f\left(\phi(p)\right) = v_T -J_\phi^\top \phi, v_T \neq 0$ is a good candidate for a time-invariant guiding vector field. The first condition can be checked with the Lyapunov function $V = \frac{1}{2}\|\phi(p)\|^2$, whose time derivative is $\dot V = \phi(p)^\top\dot\phi(p) = \phi(p)^\top J_\phi u_p = -\phi(p)^\top J_\phi J_\phi^\top \phi(p) \leq 0$, and if $p(0)\in\Omega_b$ then the equality holds if and only if $\phi = 0$ since $J_\phi J_\phi^\top$ is full rank for $p\in\Omega_b$ and $\Omega_b$ is invariant. The second condition is trivial to check since for $p\in\Omega_b$ we have that $f\left(\phi(p), t\right) \cap T_p(\mathcal{M}\phi_z) = v_T$.
\end{example}

\begin{remark}
	\label{rem: 1}
	For the special case of a path $\mathcal{P}$ in $2$D, where $\phi(p) = c \in \mathbb{R}$ is a scalar so $J_\phi^T = \nabla\phi(p)$, the guiding vector field can be easily constructed as $f\left(\phi(p)\right) = E\nabla\phi(p) - \nabla\phi(p)\phi(p)$, where $E\in\text{SO(2)}$ is a $\pm\frac{\pi}{2}$ rotation matrix.
\end{remark}

The Example \ref{ex: gvf} has shown that $J_\phi$ needs to be full rank for a particular guiding vector field and Lyapunov analysis. Let us now show this necessity more generally.

\begin{prop}
\label{prop: jinv}
Let $f: \mathbb{R}^m \times \mathbb{R} \to \mathbb{R}^m$ be a field and consider a generic path $\mathcal{P}$. The field $f$ is a guiding vector field only if the Jacobian $J_\phi$ is full rank for $p\in\Omega_b$.
\end{prop}

\begin{proof}
The claimed necessity can be quickly checked from the projection of $f$ onto $T_p(\mathcal{M}_{\phi_z})$ in order to satisfy the second condition in Definition \ref{def: gvf}, i.e.,
\begin{equation*}
    \begin{aligned}
&Proj_{T_p(\mathcal{M}_{\phi_z})}(f(\phi(p),t))\\
&:=\left(I - J_\phi(p)^\top \left(J_\phi(p) J_\phi(p)^\top \right)^{-1} J_\phi(p)\right) f(p, t),
    \end{aligned}
\end{equation*}
therefore, for the projection to exist, the inverse $\left(J_\phi(p) J_\phi(p)^\top \right)^{-1}$ needs to exist. Consequently the Jacobian $J_\phi$ needs to be full rank for $p\in\Omega_b$.
\end{proof}

However, the full rank of the Jacobian $J_\phi$ is not sufficient to have a guiding vector field since $f$ can be just orthogonal to $T_p(\mathcal{M}_{\phi_z})$ or zero regardless of the Jacobian for some $p\in\Omega_b$. In particular, we call \textit{the singularity set} to the set
\begin{equation}
\label{eq: psi}
\Psi := \{p \in \mathbb{R}^m, t\in[0,\infty): f\left(\phi(p), t\right) = 0\},
\end{equation}
	because it represents the combinations of positions and time where the vector field results in a zero velocity, meaning that the robot stops at certain time and it could get \emph{stuck}. The term \emph{singularity} comes from the common usage of normalizing the vector field in practice to guide only in direction \cite{kapitanyuk2017:TCST, de2017guidance}; hence, the associated unit vector is not well defined when $f\left(\phi(p), t\right) = 0$. Indeed, for time-invariant guiding vector fields, it is necessary to count on a full-rank $J_\phi$ at least in $\Omega_b$. That is the case for connected non-closed paths or without self-intersections since they satisfy Assumption \ref{asp:1} \cite{yao2021:Tro}. Indeed, a vector field with singularities in $\Omega_b$ is not a guiding vector field according to our Definition \ref{def: gvf}.


\begin{problem}
\label{problem}
	Create a guiding vector field for the path $\mathcal{P}$ that not only satisfies Definition \ref{def: gvf} but it also designs the transitory of the robot toward the path and its behavior around the path, i.e., to have control over the dynamics of the error signal $\phi(p(t))$ such that $\dot\phi = u_\phi$, where $u_\phi\in\mathbb{R}^{(m-1)}$ is a control input by design.
\end{problem}

\section{Inverse kinematics guiding vector field}
\label{sec: ik}

The analysis of the solution for the closed-loop system (\ref{eq: sys_phi}) is crucial, as it allows for the assessment of performance metrics such as overshoot and oscillations of the robot around $\mathcal{P}$. These metrics are essential for the stability and efficiency of the guiding system. Furthermore, the predictable behavior of the solution enhances robust navigation in complex environments by enabling the system to anticipate and effectively respond to varying conditions. According to Problem \ref{problem}, we aim to design a vector field \eqref{eq: field_xi} such that an analytical solution exists for the dynamic system \eqref{eq: sys_phi}, e.g., by the application of inverse kinematics in \eqref{eq: sys_phi} through \eqref{eq: field_xi}.

\subsection{Inverse kinematics GVF for single integrator robots}


To tame the error signal $\phi$, we seek to control its dynamics with the virtual input $u_\phi(\phi): \mathbb{R}^{m-1} \to \mathbb{R}^{m-1}$ through the following, possibly nonautonomous by design, system
\begin{equation}\label{eq:virtual_u}
    \dot{\phi} = u_\phi(\phi, t).
\end{equation}
The question now is how to design a guiding vector field $f\left(\phi(p), t\right)$ such that the error dynamics behaves as \eqref{eq:virtual_u}. We start with splitting into two components the design
\begin{equation}\label{eq:field_xi_2}
	f\left(\phi(p), t\right) = v_T + v_C, \quad v_T \in T_p(\mathcal{M}\phi_z) \setminus \{0\},
\end{equation}
where $v_C(p, t) \in \mathbb{R}^m$, the \emph{converging} velocity, will be the only responsible component of the guiding vector field to achieve \eqref{eq:virtual_u}. Considering the vector field \eqref{eq:field_xi_2} and the kinematic model \eqref{eq: sys_phi}, we know that
\begin{equation}\label{eq:sys_phi_2}
	\dot{\phi} = J_\phi(p)(v_T + v_C) =  J_\phi(p)v_C,
\end{equation}
so that we can design $v_C$ through the inverse kinematics approach, i.e., checking \eqref{eq:virtual_u} and \eqref{eq:sys_phi_2} we need to solve the following optimization problem to optimally project $u_\phi$ onto the input space, i.e., the image of $J_\phi$, for each fixed $p\in \Omega_b$
\[
\min\limits_{v_C} \|J_\phi(p) v_C - u(\phi(p),t)\|_2,
\]
whose closed-form solution is given by
\begin{equation}
	\label{eq: vc}
    v_C(\phi(p),t)= J_\phi^{\top}(p)\left[J_\phi(p)J_\phi^\top(p)\right]^{-1} u(\phi(p),t), \nonumber
\end{equation}
and since $J_\phi(p)$ is full rank in $\Omega_b$, the inverse $\left[J_\phi(p)J_\phi^\top(p)\right]^{-1}$ is well defined for $p\in\Omega_b$. Finally, taking into account $v_T$, the designed guiding vector field results in
\begin{equation}
\label{eq:IK_VF}
f\left(\phi(p), t\right) = v_T(p,t) + \underbrace{J_\phi^{\top}(p)[J_\phi(p)J_\phi^\top(p)]^{-1} u_\phi(\phi(p),t)}_{v_C(\phi(p),t)}.
\end{equation}
We call \eqref{eq:IK_VF} the \textit{inverse kinematics guiding vector field} (IK-GVF). Before giving formality to this denomination for the field, we need the following technical result.
\begin{lemma}
\label{lem: nosin}
Consider a path $\mathcal{P}$ under Assumptions \ref{asp:2} and \ref{asp:1}. If $\|v_T(p,t)\| > 0$ in $\Omega_b$ for all $t$, then \eqref{eq:IK_VF} is not singular.
\end{lemma}
\begin{proof}
We are going to show that $v_C$ and $v_T$ are orthogonal so that $v_C + v_T \neq 0$ if $v_T \neq 0$. Recall that $J_\phi v_T = 0$, then
\begin{equation*}
v_T^\top v_C = v_T^\top J_\phi^\top (J_\phi J_\phi^\top)^{-1} u_\phi = (J_\phi v_T)^\top (J_\phi J_\phi^\top)^{-1} u_\phi = 0,
\end{equation*}
since $J_\phi$ is full rank. Because $v_T \neq 0$, then \eqref{eq:IK_VF} cannot be singular in $\Omega_b$.
\end{proof}
The main difference between the IK-GVF and the \emph{classic} GVF in Example \ref{ex: gvf} is that, in the latter, we require $J_\phi$ to be full rank due to the Lyapunov analysis, whereas in the former, we need $J_\phi J_\phi^\top$ to have an inverse. In order words, the convergence rate of the level-set error signal $\phi$ in the \emph{classic} GVF depends on $J_\phi$ while for IK-GVF does not. This distinction will become clearer in the next main result, which solves Problem \ref{problem}.

\begin{theorem}
\label{thm: 1}
	Consider the vector field (\ref{eq:IK_VF}) with $\|v_T(p,t)\| > 0$ in $\Omega_b$ for all $t$. Then, the system \eqref{eq: sys_phi} is locally asymptotically stable if the system \eqref{eq:virtual_u} is globally asymptotically stable. Moreover, (\ref{eq:IK_VF}) is an inverse kinematics guiding vector field since it solves Problem \ref{problem}.
\end{theorem}

\begin{proof}
   The kinematic model given by \eqref{eq: sys_phi} with the guiding vector field \eqref{eq:IK_VF} can be expressed as
    \[
    \dot{\phi} = J_\phi J_\phi^{\top} \left[J_\phi J_\phi^\top\right]^{-1} u_\phi,
    \]
    and knowing that $J_\phi$ is full rank in $\Omega_b$, then
\begin{equation}
\label{eq: uu}
    \dot{\phi} = u_\phi, \quad p\in\Omega_b,
\end{equation}
	thus, system \eqref{eq: sys_phi} is locally stable given \eqref{eq:virtual_u} is globally asymptotically stable, e.g., $u_\phi = -\phi$. Since (\ref{eq:IK_VF}) with $\|v_T(p,t)\| > 0$ is not singular because of Lemma \ref{lem: nosin}; then, looking at (\ref{eq: uu}), we conclude that (\ref{eq:IK_VF}) is an IK-GVF that solves Problem \ref{problem}.
\end{proof}

The inverse kinematics vector field gives more flexibility to the design of the robot motion through the error signal $\phi(t)$. For example, the error system can be designed to behave as a linear one. Since the solution of linear systems are well-known analytically, we can exploit them not only to induce a desired behavior for the robot over the path but during the transitory motion towards it.

\subsection{Behavior-driven IK-GVF}

The behavior of the autonomous robot around the path $\mathcal{P}$ can be driven by a time-varying signal $\gamma(t)$, where the locally Lipschitz function $\gamma:\mathbb{R}^+\mapsto\mathbb{R}^{n-1}$ actuates over the behavior of the level set error signal $\phi(t)$. Hence, our IK-GVF will consist of three components: $v_C(p)$ to assist with the first condition of Definition \ref{def: gvf}, $v_T(p)$ to \emph{progress} along $\mathcal{P}$ as required in the second condition of Definition \ref{def: gvf}, and $v_B(p,t)$ so that $\lim_{t\to\infty}\phi(t) = \gamma(t)$, i.e., $f\left(\phi(p),t\right) = v_T(p,t) + v_C(p) + v_B(p,t)$, with 
\begin{equation}
\label{eq: vb}
J_\phi(p)(v_C(p) + v_B(p,t)) = \dot\gamma(t) - k_B(\phi(p) - \gamma(t)),
\end{equation}
where $k_B\in\mathbb{R}^+$ is a positive gain. For example, for $v_C(p) = -J_\phi^\top(p)(J_\phi(p)J_\phi^\top(p))^{-1}k_B\phi(p)$, we can choose $v_B(p,t) = J_\phi^\top(p)(J_\phi(p)J_\phi^\top(p))^{-1}(\dot\gamma(t) + k_B \gamma(t))$.

\section{Inverse kinematic guiding vector filed for a fixed-wing UAV}
\label{sec: uni}

In this section, we demonstrate how to design a behavior-driven IK-GVF for a fixed-wing travelling with constant speed whose dynamics are
\begin{equation}
	\label{eq: uni}
    \left\{
    \begin{aligned}
        \dot{p} &= v R(\theta) p_0,\\
        \dot{\theta} &= \omega,
    \end{aligned}
    \right.
\end{equation}
where $p = [p_x,p_y]^\top\in \mathbb{R}^2$ is the \emph{horizontal} UAV position, $v\in \mathbb{R}^+$ is a constant speed, $R(\theta) = \left[\begin{smallmatrix}
    \cos\theta & -\sin \theta\\
    \sin\theta & \cos\theta
\end{smallmatrix}\right] = e^{E\theta}$ with $E = \left[\begin{smallmatrix}
    0 & -1\\
    1 & 0
\end{smallmatrix}\right]$ represents the \emph{heading} of the vehicle, $p_0 = \left[\begin{smallmatrix}
    1\\
    0
\end{smallmatrix}\right]$, and $\omega\in\mathbb{R}$ is the actuation or \emph{heading rate}.

\begin{remark}
The design of an IK-GVF for a fixed-wing drone demands an extra requirement for $f$ to be used with the dynamics (\ref{eq: uni}), i.e., the IK-GVF must be sufficiently smooth such that the required heading rate to track $f$ is bounded.
\end{remark}
	Let us consider the following template
\begin{equation}
	\label{eq:xi_unicycle}
	f(\phi(p), t) = \alpha \,\hat{v}_T(p) + \vartheta(p,t),
\end{equation}
where $\hat{v}_T(p) = \frac{v_T(p)}{\|v_T(p)\|}$, $\vartheta(p,t) = v_C(p) + v_B(p,t)$ with $v_C$ and $v_B$ satisfying (\ref{eq: vb}), and $\alpha$ will be a function that will assist us to keep $\|f(\phi(p), t)\| = v$. Note that we have dropped $t$ from $v_C$. The idea is to design the IK-GVF such that the guiding field has the same norm as the speed of the UAV; hence, if the vehicle is perfectly aligned to the field, then we can precisely control the dynamics and transitory of the level-set error signal $\phi(t)$. In particular, if\footnote{For the sake of convinience, allow us to relax Definition \ref{def: gvf}. For example, $\|\vartheta(p,t)\| = v$ if and only if $p\in\partial\Omega_b$, i.e, the vehicle will not \emph{progress} along the path when $p\in\partial\Omega_b$ but still we will call (\ref{eq:xi_unicycle}) an IK-GVF in $\Omega_b$.} $\|\vartheta(p,t)\| \leq v$, then we need to add $v_T$ weighted by $\alpha$ appropriately in order to have $\|f(\phi(p), t)\| = v$. For example, for a time-invariant IK-GVF, we can design $v_C$ such that there is a compact $\Omega_c$ where $\|v_C \| \leq v$, e.g., as in Example \ref{ex: gvf}, and if $v_B$ is bounded then we can also find $\Omega_b\subseteq\Omega_c$ such that $\|v_C + v_B\| \leq v$. In particular, we choose $\alpha$ as follows
\begin{align}
	(\alpha \, \hat{v}_T + \vartheta)^\top (\alpha \, \hat{v}_T + \vartheta) &= v^2 \nonumber \\
	\alpha^2 \, \|\hat{v}_T\|^2 + \|\vartheta\|^2 + 2\alpha \, \hat{v}_T^\top\vartheta &= v^2 \nonumber \\
	\alpha^2 +  \|\vartheta\|^2 = v^2 \implies &\alpha = \sqrt{v^2 - \|\vartheta\|^2}  \nonumber,
\end{align}
where we note that $\hat{v}_T^\top\vartheta = 0$ since the two vectors are orthogonal. When $p\notin\Omega_b$, then we have that $\|v_C + v_B\| > v$; hence, we propose the following vector field that is an IK-GVF within $\Omega_b$
\begin{equation}
	\label{eq: gvfuni}
	f\left(\phi(p),t\right)  =\left\{
    \begin{aligned}
        &  v\tfrac{\vartheta}{\|\vartheta\|}, \  \|\vartheta\| > v\\
        & \alpha \, \hat{v}_T + \vartheta, \ \|\vartheta\| \le v
    \end{aligned}\right. ,
\end{equation}
where it is straightforward to see that (\ref{eq: gvfuni}) is continuous. Specifically, when $|\vartheta| = v$, then $f\left(\phi(p), t\right) = v \frac{\vartheta}{\|\vartheta\|}$ since $\alpha = 0$, which is also the trivial limit of $f$ as $\|\vartheta\| \to v$. Additionally, $J\phi(p)$ must be well-defined or different from zero in the set $\Omega_c \supseteq \Omega_b$ for $\frac{\vartheta}{\|\vartheta\|}$ to be well-defined in $\Omega_c$.
\begin{remark}
Note that the field (\ref{eq: gvfuni}) has velocity units, e.g., m/s; allowing a straightforward physical interpretation of the aircraft behavior.
\end{remark}
If $p(0) \notin \Omega_b$ and the robot is perfectly aligned with the field (\ref{eq: gvfuni}), then $p(t)$ will go straight to enter the invariant set $\Omega_b$ in finite time, where the error signal $\phi$ behaves as in \eqref{eq: vb}. If the vehicle is not aligned, an additional controller is needed to align the robot's heading with $f$, in addition to the desired angular velocity $\dot{\theta}_c \in \mathbb{R}$ derived from $f$ to track it.


\begin{theorem}
\label{thm: 2}
	Assume that $J_\phi(p)$ is full rank in $\Omega_c \supseteq \Omega_b$, where $||\vartheta|| \leq v$ only for $p\in\Omega_b$, and $p(0)\in\ \left\{\Omega_c \, \setminus \, \partial\Omega_c\right\}$. Consider for (\ref{eq: uni}) the controller
    \begin{equation}\label{eq:omega}
        \omega = \dot{\theta
    }_c + k_\theta\frac{f^\top E \dot{p}}{v^2} ,
    \end{equation}
where $k_\theta\in\mathbb{R}^+$, $f$ is as in (\ref{eq: gvfuni}) such that $f$ and $\dot f$ are sufficiently smooth in $\Omega_c$, and $\dot{\theta}_c = \frac{1}{v^2} f^\top E^\top \dot{f}$. Then, $
	\|f\left(\phi(p), t\right)-\dot{p}(t)\|\rightarrow 0$ as $t\rightarrow \infty$.
\end{theorem}
\begin{proof}
The sketch of the proof is as follows. Since $f\left(\phi(p),t\right)$ and $\dot{p}$ has the same norm $v$, we can define the orientation error as $\varphi = \dot{p} - f$, whose dynamics $\dot\varphi = \ddot p - \dot f$ are given by
\begin{equation}\label{eq:varphi}
    \dot{\varphi} = \dot \theta E\dot{p} - \dot{\theta}_c Ef,
\end{equation}
where the derivation of $\dot f$ can be done from \cite[Lemma 3.1]{kapitaniuk2020motion}. Consider the following Lyapunov function candidate $V(t) = \tfrac{1}{2}\|\varphi(t)\|^2$, whose time derivative along \eqref{eq:varphi} is
\begin{equation}
    \begin{aligned}
        \dot{V} = (\dot{p}-f)^\top(\ddot{p}-\dot{f}) = -\dot{\theta}_c\dot{p}^\top E f - \dot{\theta}f^\top E \dot{p}.
    \end{aligned}
\end{equation}
	Knowing that $E^\top = -E$, and using the controller \eqref{eq:omega}, we conclude that $\dot{V} = -k_\theta\frac{(f^\top E\dot{p})^2}{v^2} \leq 0,$ where if $\dot p(0)$ and $f$ are not forming $\pi$ radians, then the equality holds if and only if $\varphi = 0$ since $J_\phi$ is full rank in $\Omega_c$ and it is invariant if $p(0)$ is inside $\Omega_c$ but not at its boundary counting on a $k_\theta$ sufficiently big, i.e., the vehicle has enough time to align its heading with the guiding vector field $f$ before possibly leaving $\Omega_c$. Hence, the origin of \eqref{eq:varphi} is asymptotically stable.
\end{proof}

For the controller (\ref{eq:omega}) we provide the expressions
\begin{equation}
    \begin{aligned}
        \dot{f}  =& \left\{
    	\begin{aligned}
		& v \left(\tfrac{I_2}{\|\vartheta\|} - \tfrac{\vartheta \vartheta^\top}{\|\vartheta\|^3}\right) \dot{\vartheta}, \  \|\vartheta\| > v\\
		& \dot{\alpha} \tfrac{v_T(p)}{\|v_T(p)\|}+ \alpha \left(\tfrac{I_2}{\|v_T\|} - \tfrac{v_T v_T^\top}{\|v_T\|^3}\right) \dot{v}_T + \dot{\vartheta}, \ \|\vartheta\| \le v
    	\end{aligned}\right. ,\\
        \dot{\vartheta} =& \dot{\beta} (u_\phi + \dot\gamma(t))
        	+ \beta (\dot{u}_\phi+ \ddot{\gamma}(t)),\\
        \dot{\beta} =& \tfrac{H_{\phi}^\top \dot{p} - 
        	J_{\phi}^\top (J_{\phi} J_{\phi}^\top)^{-1} \left[\dot{p}^\top H_{\phi} J_{\phi}^\top + J_{\phi} H_{\phi}^\top f\right] }
        	{J_{\phi} J_{\phi}^\top}, \nonumber
    \end{aligned}
\end{equation}
where $I_2$ is the $2\times 2$ identity matrix, $\beta = J_{\phi}^\top (J_{\phi} J_ {\phi}^\top)^{-1}$, and $H_{\phi(p)} \in \mathbb{R}^{2 \times 2}$ is the Hessian of the level set $\phi(p)$ at $p$. Note that for $\dot f$, the expression $v \left(\tfrac{I_2}{\|\vartheta\|} - \tfrac{\vartheta \vartheta^\top}{\|\vartheta\|^3}\right) \dot{\vartheta} = 0$ since it is the orthogonal projection of $\dot\vartheta$ onto $\vartheta$, and since $v_C$ lives in a space that has dimension $1$ because of the considered $2$D Euclidean space, $\dot\vartheta$ and $\vartheta$ must be parallel.

\section{Flight experiments}
\label{sec: sim}
\begin{figure}
\centering
\includegraphics[width=1\columnwidth]{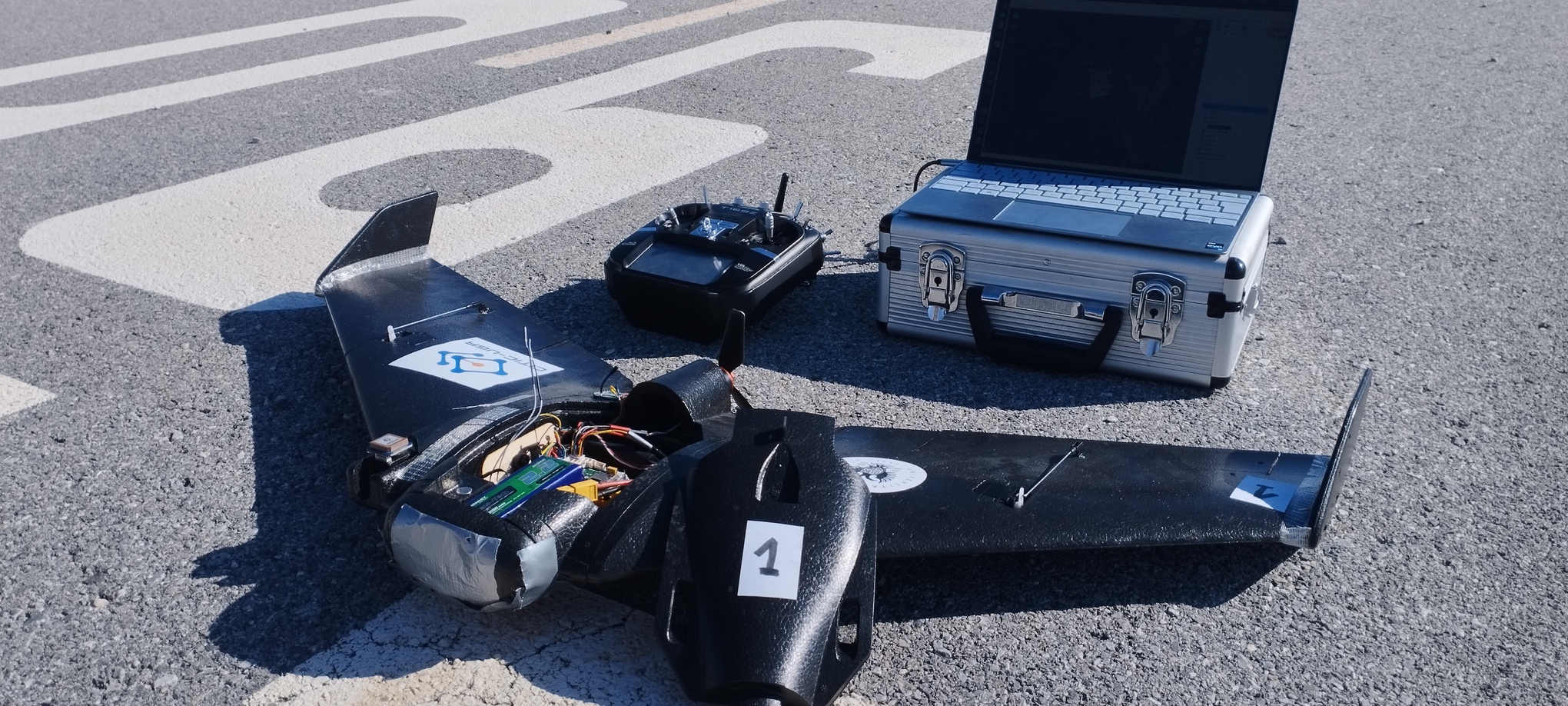}
\caption{0.9m wingspan fixed-wing drone used during the experiments.}
\label{fig: sonic}
\end{figure}

We validate Theorems \ref{thm: 1} and \ref{thm: 2} in different scenarios with a fixed-wing drone. The aircraft in Figure \ref{fig: sonic} is a Sonicmodell AR.Wing made of EPP with 0.9m wingspan equipped with a 2200kv motor, two 9gr servos, Ublox GPS, 3S 2200mAh LiPo battery, Futaba transmitter for safe manual radio-control if necessary, Xbee Pro S1 for telemetry, and an Apogee board running Paparazzi firmware \cite{brisset2006paparazzi, pprzurl} in a microcontroller\footnote{At the moment of the submission, the code can be found at \url{https://github.com/Swarm-Systems-Lab/gvf_ik/tree/master}}. The total weight is 650gr with an autonomy of 25 minutes flying at 50 meters above ground (around 720m above mean sea level) with an estimated airspeed of 14m/s. The ground station is a standard laptop and another Xbee Pro S1 with a dipole antenna running Paparazzi software. The flights took place at the club \emph{Ciudad de la Alhambra}\footnote{Coordinates: 37°17'51.8"N, 3°40'58.4"W.} in Granada, Spain.

\subsection{Controlled exponential convergence to a circumference}
\begin{figure*}[t]
\centering
\includegraphics[width=2\columnwidth]{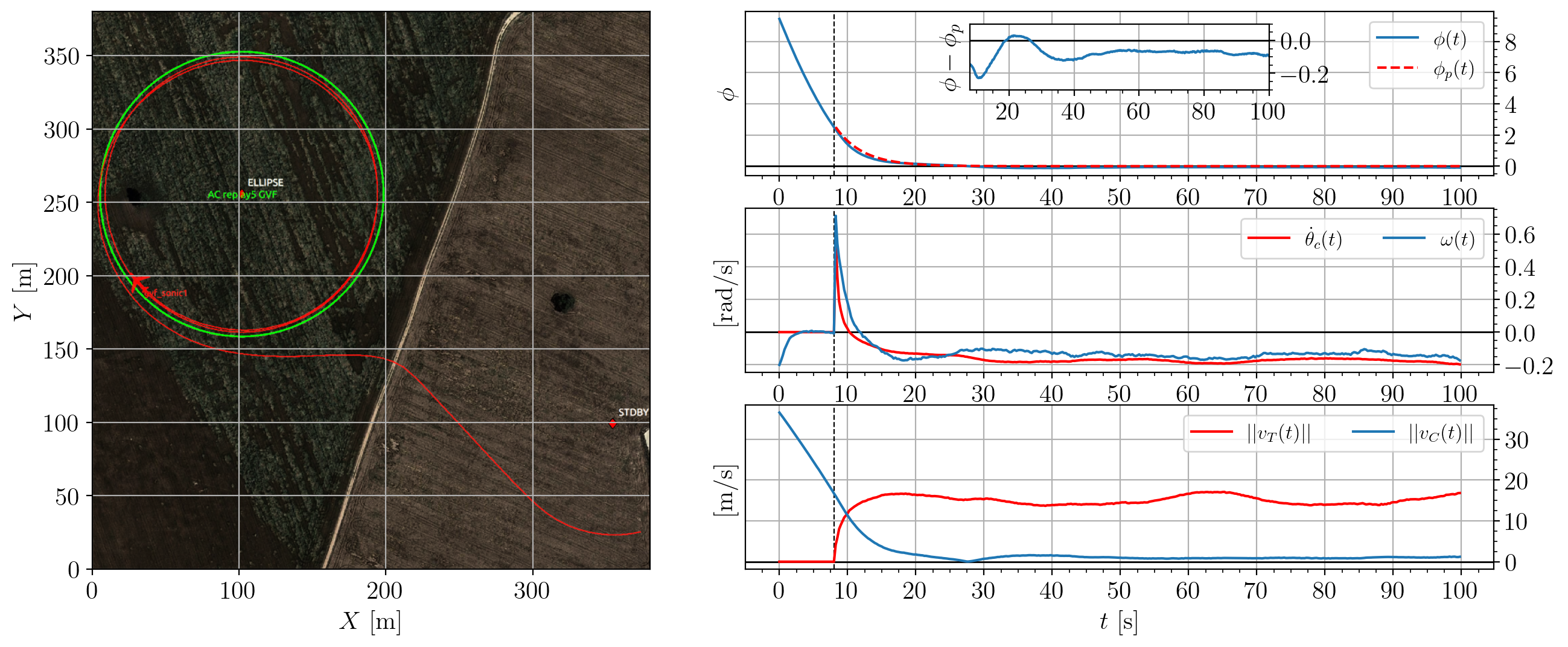}
\caption{
The fixed-wing drone in Figure \ref{fig: sonic}, following the IK-GVF \eqref{eq:IK_VF}, converges to a circular path with radius $r=100$ meters. The ground speed of the drone is between 14 and 17 m/s, depending on the wind (around 2 m/s). The design parameters are $k_\phi = 0.25$ and $k_\theta = 0.6$. On the right, from top to bottom, the actual kinematic of $\phi(t)$ in comparison with the predicted one, given by $\dot \phi_p(t) = - k_\phi \phi_d(t)$, starting from $t_p = 8$ s (vertical dashed black line). Prior to this, the drone aligns with $f$ and proceeds directly toward the path. The drone tracks the designed transitory satisfactorily as predicted. Next is the control input $\omega$ commanded by the controller (as defined in \eqref{eq:omega}), along with $\dot \theta_c = \frac{1}{v^2}f^\top E^\top \dot f$. Finally, the norms of $v_T$ and $v_C$, which oscillate due to the wind, are displayed.}
\label{fig: ex1}
\end{figure*}

In this flight, the path to be followed is a circumference with the normalized implicit equation $\phi(p) := \frac{1}{r^2} \left((p_x - x_0)^2 + (p_y-y_0)^2\right) - 1$, where $r\in\mathbb R^+$ is the radius and $\begin{bmatrix}x_0 & y_0\end{bmatrix}^\top$ is the center. The Jacobian is $J_{\phi(p)} = \nabla\phi(p)^\top = \frac{1}{r^2}\begin{bmatrix}2(p_x - x_0) & (p_y - y_0)\end{bmatrix}$ and Hessian $H_{\phi(p)} = \frac{1}{r^2}\left[\begin{smallmatrix}2 & 0 \\ 0 & 2\end{smallmatrix}\right]$. For the inverse kinematics guiding vector field (\ref{eq: gvfuni}) we choose $v_T = EJ_\phi^\top = E\nabla\phi(p)$ and $\vartheta = v_C$ with $v_C$ as in (\ref{eq:IK_VF}) choosing $u_\phi = -k_\phi \phi$ so that the level-set error signal decreases exponentially fast with time constant $k_\phi$ once $\|u_\phi\| \leq v$. Since the ground speed varies due to the presence of wind, in the onboard software we set $v$ to the actual GPS speed. We show the flight results in Figure \ref{fig: ex1}. Since $\phi(p)$ is normalized, the physical interpretation of the signal is as follows. In polar coordinates, i.e., $p = \rho\begin{bmatrix}\cos\theta & \sin
\theta\end{bmatrix}^\top$, where $\rho \in \mathbb{R}^+$ and $\theta \in \mathcal{S}^1$, centered at $\begin{bmatrix}x_0 & y_0\end{bmatrix}^\top$, we have that $\phi(p) = \frac{\rho^2}{r^2} - 1$. Therefore, the equivalent error in distance units is $r \sqrt{\phi + 1} - r$, e.g, an error of $\phi = 0.1$ corresponds to a radial error of $4.88$m.

\subsection{Improving transmission signal}
\begin{figure}[h!]
\centering
\includegraphics[width=1\columnwidth]{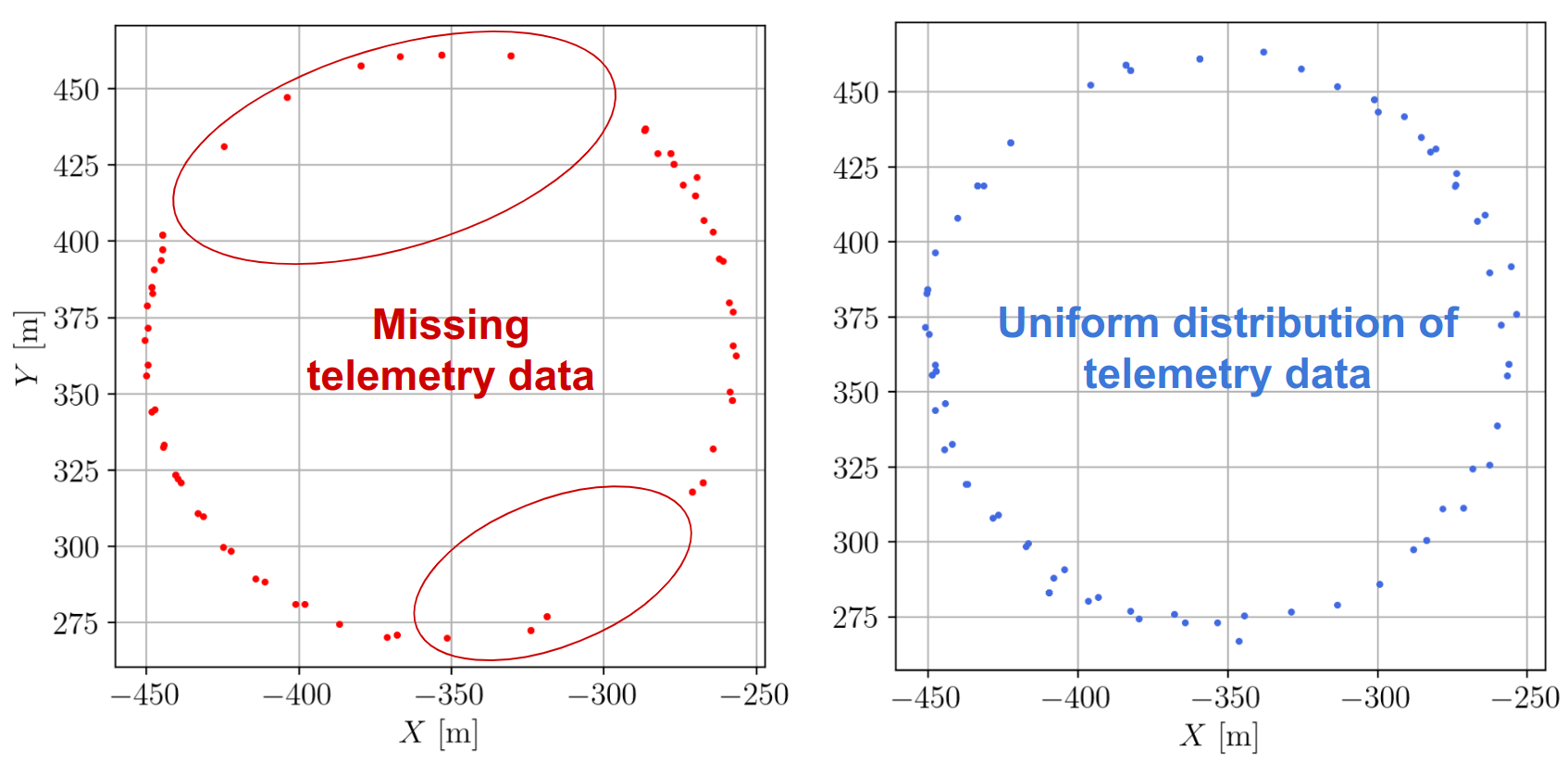}
\caption{
Telemetry data of the X-Y position received in the ground station at a frequency of 1Hz from the fixed-wing drone, which tracks two different IK-GVFs to follow a circular path of radius $r = 100$ meters. On the left (red dots), the desired kinematic of the error signal follows $\dot \phi = - k_\phi \phi(t)$, while on the right (blue dots) $\dot \phi = \dot \gamma(t) - k_\phi (\phi(t) - \gamma(t))$, where $\gamma(t) = A \sin(\omega_\gamma t)$, with $A = 0.03$ and $\omega_\gamma = 1.3$ rad/s, drives the desired behavior. For both vector fields, the design parameters are $k_\phi = 0.25$ and $k_\theta = 0.8$. Transmission quality improves (resulting in more uniformly distributed data) when the controlled oscillations are induced.}
\label{fig: ex3}
\end{figure}

\begin{figure*}[t]
\centering
\includegraphics[width=2\columnwidth]{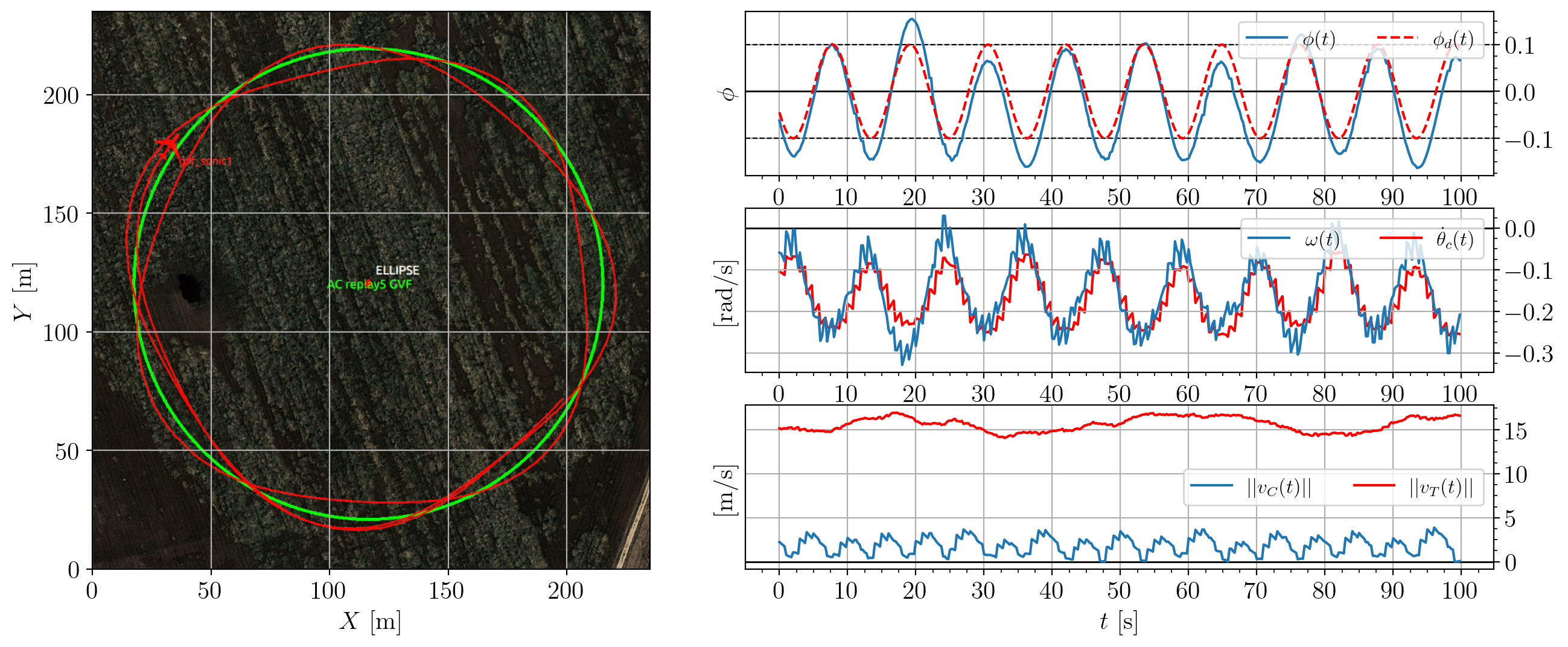}
\caption{
The fixed-wing drone in Figure \ref{fig: sonic} converges to a circular path of radius $r = 100$ by following a behavior-driven IK-GVF with $\gamma(t) = A \sin(\omega_\gamma t)$, where $A = 0.1$, and $\omega_\gamma = 0.55$ rad/s. The ground speed of the drone is between 14 and 17 m/s, depending on the wind (around 2m/s). The design parameters are $k_\phi = 0.25$ and $k_\theta = 0.8$. On the right, from top to bottom, the actual kinematic of $\phi(t)$ in comparison with the predicted one, given by $\dot \phi_p = \dot \gamma(t) - k_\phi (\phi(t) - \gamma(t))$, where $\gamma(t) = A \sin(\omega_\gamma t)$ drives the desired behavior. Regarding the initial condition, $\phi_p(0) = \phi(0) + \gamma(t_p)$, with $t_p = 6.55$ s. The phase between the red and blue signals have been adjusted by hand since we are inducing $\dot\gamma$ and not $\gamma$. The error between the actual and desired behavior is around 5 meters. Next is the control input $\omega$, commanded by the controller (as defined in \eqref{eq:omega}), along with $\dot \theta_c = \frac{1}{v^2}f^\top E^\top \dot f$. Finally, the norms of $v_T$ and $v_C$, which oscillate due to the wind, are displayed.}
\label{fig: ex2}
\end{figure*}

For this experiment we induce an oscillatory motion with $\gamma = A\sin(\omega_\gamma t)$ for (\ref{eq: vb}) for the previous circular path. We show two flights. The first flight, shown in Figure \ref{fig: ex2}, happens close to the ground station to capture all the telemetry. The second flight happens far from the ground station where telemetry reception starts to fail. Thanks to the oscillatory motion that affects the roll angle (see the attached video) we can see in Figure \ref{fig: ex3} how the telemetry points are better uniformly distributed than when no oscillation is induced.

\section{Conclusions}
\label{sec: con}
We have analyzed and applied the concept of inverse kinematics to \emph{classic} guiding vector fields. The proposed technique allows us to precisely control the transitory of the robot toward the desired path plus inducing a desired behavior around it with guarantees. We have showcased the effectiveness of the technique with fixed-wing drones even improving their telemetry reception. In addition we provide the source code of the implementation in the open-source project for autonomous vehicles \emph{Paparazzi}.

\bibliographystyle{IEEEtran}
\bibliography{reference}

\begin{thebibliography}{10}
\providecommand{\url}[1]{#1}
\csname url@samestyle\endcsname
\providecommand{\newblock}{\relax}
\providecommand{\bibinfo}[2]{#2}
\providecommand{\BIBentrySTDinterwordspacing}{\spaceskip=0pt\relax}
\providecommand{\BIBentryALTinterwordstretchfactor}{4}
\providecommand{\BIBentryALTinterwordspacing}{\spaceskip=\fontdimen2\font plus
\BIBentryALTinterwordstretchfactor\fontdimen3\font minus
  \fontdimen4\font\relax}
\providecommand{\BIBforeignlanguage}[2]{{%
\expandafter\ifx\csname l@#1\endcsname\relax
\typeout{** WARNING: IEEEtran.bst: No hyphenation pattern has been}%
\typeout{** loaded for the language `#1'. Using the pattern for}%
\typeout{** the default language instead.}%
\else
\language=\csname l@#1\endcsname
\fi
#2}}
\providecommand{\BIBdecl}{\relax}
\BIBdecl

\bibitem{paul1981robot}
R.~P. Paul, \emph{Robot manipulators: mathematics, programming, and control:
  the computer control of robot manipulators}.\hskip 1em plus 0.5em minus
  0.4em\relax Richard Paul, 1981.

\bibitem{choset2005principles}
H.~Choset, K.~M. Lynch, S.~Hutchinson, G.~A. Kantor, and W.~Burgard,
  \emph{Principles of robot motion: theory, algorithms, and
  implementations}.\hskip 1em plus 0.5em minus 0.4em\relax MIT press, 2005.

\bibitem{yao2021:Tro}
W.~Yao, H.~G. de~Marina, B.~Lin, and M.~Cao, ``Singularity-free guiding vector
  field for robot navigation,'' \emph{IEEE Transactions on Robotics}, vol.~37,
  no.~4, pp. 1206--1221, 2021.

\bibitem{kapitanyuk2017:TCST}
Y.~A. Kapitanyuk, A.~V. Proskurnikov, and M.~Cao, ``A guiding vector-field
  algorithm for path-following control of nonholonomic mobile robots,''
  \emph{IEEE Transactions on Control Systems Technology}, vol.~26, no.~4, pp.
  1372--1385, 2017.

\bibitem{lawrence2007:GNCC}
D.~Lawrence, E.~Frew, and W.~Pisano, ``Lyapunov vector fields for autonomous
  {UAV} flight control,'' in \emph{AIAA Guidance, Navigation and Control
  Conference and Exhibit}, 2007, p. 6317.

\bibitem{goncalves2010:Tro}
V.~M. Goncalves, L.~C. Pimenta, C.~A. Maia, B.~C. Dutra, and G.~A. Pereira,
  ``Vector fields for robot navigation along time-varying curves in $ n
  $-dimensions,'' \emph{IEEE Transactions on Robotics}, vol.~26, no.~4, pp.
  647--659, 2010.

\bibitem{de2017guidance}
H.~G. De~Marina, Y.~A. Kapitanyuk, M.~Bronz, G.~Hattenberger, and M.~Cao,
  ``Guidance algorithm for smooth trajectory tracking of a fixed wing {UAV}
  flying in wind flows,'' in \emph{2017 IEEE international conference on
  robotics and automation (ICRA)}.\hskip 1em plus 0.5em minus 0.4em\relax IEEE,
  2017, pp. 5740--5745.

\bibitem{golubitsky2012:book}
M.~Golubitsky and V.~Guillemin, \emph{Stable mappings and their
  singularities}.\hskip 1em plus 0.5em minus 0.4em\relax Springer Science \&
  Business Media, 2012, vol.~14.

\bibitem{khalil}
H.~Khalil, \emph{Nonlinear systems}.\hskip 1em plus 0.5em minus 0.4em\relax
  Englewood Cliffs, NJ: Prentice-Hall, 1996.

\bibitem{do2016:book}
M.~P. Do~Carmo, \emph{Differential geometry of curves and surfaces: revised and
  updated second edition}.\hskip 1em plus 0.5em minus 0.4em\relax Courier Dover
  Publications, 2016.

\bibitem{lee2012:book}
J.~M. Lee, \emph{Smooth manifolds}.\hskip 1em plus 0.5em minus 0.4em\relax
  Springer, 2012.

\bibitem{kapitaniuk2020motion}
Y.~Kapitaniuk, ``\BIBforeignlanguage{English}{Motion control algorithms for
  mobile vehicles and marine crafts},'' Ph.D. dissertation, University of
  Groningen, 2020.

\bibitem{brisset2006paparazzi}
P.~Brisset, A.~Drouin, M.~Gorraz, P.-S. Huard, and J.~Tyler, ``The paparazzi
  solution,'' in \emph{MAV 2006, 2nd US-European competition and workshop on
  micro air vehicles}, 2006.

\bibitem{pprzurl}
Paparazzi, ``Paparazzi open-source project,''
  \url{https://github.com/paparazzi/paparazzi}, 2006, accessed: 18-02-2025.

\end{thebibliography}


\end{document}